\documentclass[runningheads]{llncs}

\usepackage{eccv}

\usepackage{eccvabbrv}

\usepackage{graphicx}
\usepackage{amsmath}
\usepackage{amssymb}
\usepackage{booktabs}
\usepackage{paralist}
\usepackage{bm}

\usepackage{xcolor}
\usepackage{booktabs}       
\usepackage{amsfonts}       
\usepackage{nicefrac}       
\usepackage{microtype}      
\usepackage{xcolor}         
\usepackage{graphicx}
\usepackage{amsmath}
\usepackage{amssymb}
\usepackage{amsfonts}
\usepackage{multirow}
\usepackage{color}
\usepackage{colortbl}
\usepackage{wrapfig}
\usepackage{balance}
\usepackage{amsmath}
\usepackage{amssymb}
\usepackage{xcolor}
\usepackage{makecell}

\usepackage{adjustbox}
\usepackage{comment}

\usepackage{booktabs}
\usepackage{multirow}
\usepackage{makecell}
\usepackage{caption} 
\usepackage[symbol]{footmisc}

\usepackage[accsupp]{axessibility}  

\usepackage{hyperref}

\usepackage{orcidlink}

\begin{document}

\title{WTS: A Pedestrian-Centric Traffic Video Dataset for Fine-grained Spatial-Temporal Understanding} 

\titlerunning{WTS: A Pedestrian-Centric Traffic Video Dataset}

\author{Quan Kong\inst{1}\thanks{quan.kong@woven.toyota} \and
Yuki Kawana\inst{1} \and
Rajat Saini \inst{1} \and
Ashutosh Kumar \inst{1} \and
Jingjing Pan \inst{1} \and
Ta Gu \inst{2}\thanks{Work done while Ta Gu was an intern at Woven by Toyota.} \and
Yohei Ozao \inst{1} \and
Balazs Opra \inst{1} \and
David C. Anastasiu \inst{3} \and
Yoichi Sato \inst{2} \and
Norimasa Kobori \inst{1}
}

\authorrunning{Q.Kong et al.}

\institute{$^1$Woven by Toyota, $^2$The University of Tokyo, $^3$Santa Clara University
}

\maketitle

\begin{abstract}
In this paper, we address the challenge of fine-grained video event understanding in traffic scenarios, vital for autonomous driving and safety. Traditional datasets focus on driver or vehicle behavior, often neglecting pedestrian perspectives. To fill this gap, we introduce the WTS dataset, highlighting detailed behaviors of both vehicles and pedestrians across over $1.2k$ video events in over hundreds traffic scenarios. WTS integrates diverse perspectives from vehicle ego and fixed overhead cameras in a vehicle-infrastructure cooperative environment, enriched with comprehensive textual descriptions and unique 3D Gaze data for a synchronized 2D/3D view, focusing on pedestrian analysis. We also provide annotations for $5k$ publicly sourced pedestrian-related traffic videos. Additionally, we introduce LLMScorer, an LLM-based evaluation metric to align inference captions with ground truth. Using WTS, we establish a benchmark for dense video-to-text tasks, exploring state-of-the-art Vision-Language Models with an instance-aware VideoLLM method as a baseline. WTS aims to advance fine-grained video event understanding, enhancing traffic safety and autonomous driving development. Dataset page: \url{https://woven-visionai.github.io/wts-dataset-homepage/}.
\end{abstract}

\section{Introduction}
\label{sec:intro}

\begin{figure*}[t]
\centering
\includegraphics[scale=0.48]{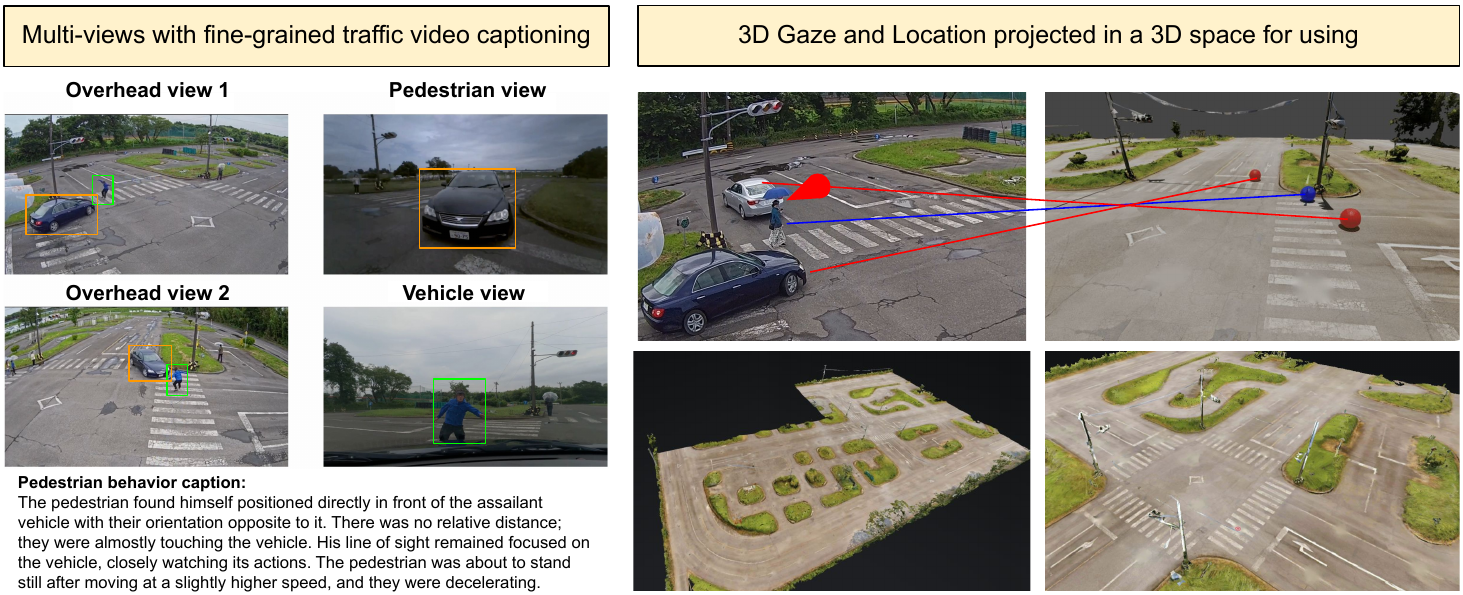}
   \caption{The overview of WTS dataset features. We provide multi-view videos with fine-grained video captions focusing on pedestrian behavior and the 3D gaze and location information for a further detailed understanding of the traffic-related videos.}
\label{fig:intro}
\end{figure*}

Understanding fine-grained information from videos has been a paramount challenge in computer vision, especially in mission-critical applications like autonomous driving and traffic safety scenario analysis \cite{xu2023drivegpt4,sima2023drivelm,MallaCDCL23}. 
This challenge hinges on interpreting complex spatial-temporal data swiftly and accurately, encompassing environmental context and individual behaviors for robust decision-making and causal understanding of user intentions. Despite significant advancements in this domain, several gaps persist, which we aim to address in our work.

Existing research extensively focuses on vehicle and driver behavior, but pedestrian behavior—a critical aspect of traffic safety—remains underexplored, despite statistics showing over $20\%$ \cite{traffic_statiscal} of traffic accidents involve pedestrians. Current traffic event models lack granularity in behavior definition, limiting nuanced decision-making. The rise of Multi-modal Large Language Models (MLLMs), which integrate large language models with multi-modality, advances the generation of detailed textual descriptions from images or videos. However, applying MLLMs to interpret fine-grained, domain-specific details within traffic scenarios is still challenging and underdeveloped for real-world applications.

To address these gaps, we propose the WTS dataset, a pedestrian-centric traffic video dataset with detailed textual descriptions of both pedestrian and vehicle behaviors. We recorded traffic scenarios from multiple views, using overhead and vehicle drive recorder cameras, with five video segment labels for traffic accident analysis. The dataset includes $1.2K$ well-annotated dense descriptions across $255$ traffic scenarios. Multiple-view videos are synchronized using an audio signal from a radio on the same channel attached to each camera. Additionally, we provide a 2D/3D synchronized space for our recording environment, offering accurate 3D gaze annotation data of pedestrians using Tobii Pro Glasses 3. The dataset includes $132$ traffic accident patterns described using the ISO34502 standard, with videos in high $1080p$ resolution at $24$ fps. Figure \ref{fig:intro} provides an overview of our dataset features. For broader experimental purposes, we also offer detailed textual descriptions for approximately $5K$ publicly sourced pedestrian-related traffic videos.

\begin{figure}[t]
\centering
\includegraphics[scale=0.30]{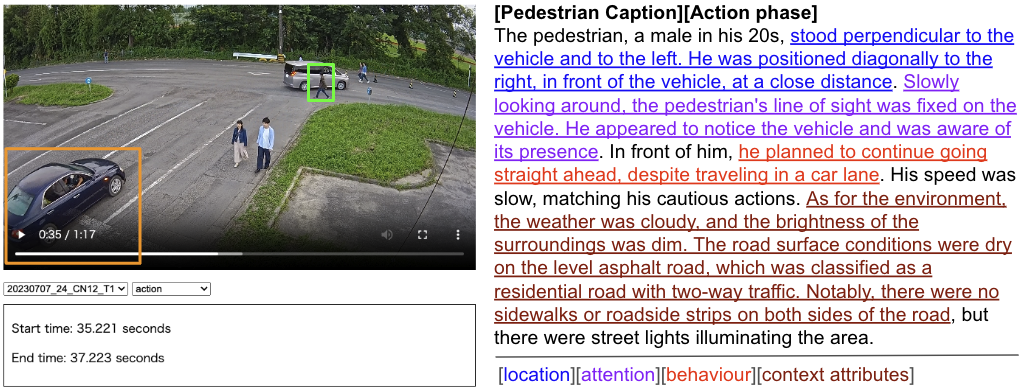}
   \caption{A full caption example with its structure design for the [action] phase}
\label{fig:caption_structure}
\end{figure}

As a benchmark for fine-grained video-to-text tasks using the WTS dataset, captions cover four high-level categories: \textit{Location, Attention, Behavior, and Context}, each with detailed textual information. The average caption length for one video segment is about 58.7 words. Figure \ref{fig:caption_structure} shows a full caption example. Traditional metrics for video/image caption evaluation, which use text embedding similarity \cite{lin-2004-rouge,papineni-etal-2002-bleu,oliveira-dos-santos-etal-2021-cider}, struggle with long descriptions as they measure word-level rather than semantic similarity between inference and ground truth sentences. To address this, we propose an LLM-based video caption scorer focusing on semantic similarity. Additionally, we introduce an instance-aware approach based on the Video Large Language Model, serving as a baseline for the fine-grained video-to-text challenge in WTS.

As a summarization of our contribution to the field in several significant ways:

\textbf{A novel pedestrian-centric traffic video dataset:} we introduce a unique dataset focusing on pedestrian-related traffic scenarios. Each traffic event in this dataset is accompanied by detailed textual descriptions of both vehicle and pedestrian behavior, annotated with structured knowledge from traffic safety analysts. 3D Gaze of the pedestrian as a meta-analysis factor in traffic safety is also provided.

\textbf{An LLM-based video caption evaluation scorer:} we introduce new metrics cards composited LLM-based scorer for better alignment with evaluating the semantic correctness than only word-level similarity.   

\textbf{Empirical Evaluation with Vision-Language Models:} to demonstrate the efficacy of our dataset, we conduct extensive experiments using cutting-edge vision-language models, including a proposed instance-aware VideoLLM. 
\section{Related Works}
\label{sec:relatedwork}

\def\arraystretch{1.3}
\begin{table}[t]
\centering
\begin{adjustbox}{width=\textwidth}
\begin{tabular}
{|l |c |c |c |c |c |c |c |c |c|}
\hline
Datasets & Videos (total) & Type & Domain & Captions num. & \thead{Avg.\\caption len.} & Year\\
\hline
\hline
MSVD \cite{chen-dolan-2011-collecting} & 1,970 & scene & open & 80,380 & 7.14 & 2011  \\
\hline
TACoS \cite{regneri-etal-2013-grounding}& 7,206  & scene & cooking & 18,227 & 8.27 & 2013 \\
\hline
MPII-MD \cite{rohrbach15cvpr}& 68,327 &scene & movie & 68,375 & 11.05 & 2015  \\
\hline
M-VAD \cite{pini2019mvad}& 46,589 & scene & movie & 46,589 & 12.44 & 2015  \\
\hline
MSR-VTT \cite{xu2016msr}& 507,502 & scene & open & 200,000 & 9.27 & 2016  \\
\hline
Charades \cite{sigurdsson2016hollywood}& 9,848 & scene & daily indoor & 25,032 & 23.91 & 2015\\
\hline
Charades-Ego \cite{sigurdsson2018actor}& 7,860 & scene & daily indoor & 14,039 & 26.30 & 2016 \\
\hline
TGIF \cite{li2016tgif}& 125,782 & scene & open & 125,781 & 11.28 & 2016 \\
\hline
ActivityNet Caps. \cite{caba2015activitynet}& 19,994 & instance & human activity & 72,976 & 14.72 & 2017 \\
\hline
VATEX \cite{wang2019vatex}& 34,991 & scene & open & 349,910 & 15.25 & 2019 \\
\hline
HowTo100M \cite{miech19howto100m}& 139,668,840 & scene & instruction & 139,668,840 & 4.16 & 2019 \\
\hline
TRECVID-VTT'20 \cite{awad2021trecvid}& 9,185 & scene & open & 28,183 & 18.90 & 2020 \\
\hline
\hline
BDD-X \cite{kim2018textual}& 6,984 & scene & traffic + outdoor & 26,228 & 14.5 & 2018  \\
\hline
\textbf{WTS} & \textbf{6,061 (1,200+4,861)} & \textbf{instance} & \textbf{traffic + outdoor} & \textbf{49,860} & \textbf{58.7} & \textbf{2023} \\
\hline

\end{tabular}
\end{adjustbox}
\caption{Comparison between different video caption and 3D gaze-related datasets.}
\label{tbl:datasets_overview}
\end{table}

In the evolution of video captioning and behavior-understanding datasets, a significant focus has been placed on varying domains and the granularity of annotations. Our dataset, WTS, stands out in its comprehensive coverage of traffic scenarios with a pedestrian-centric focus. We now draw comparisons with other datasets to highlight WTS's unique contributions to the field.

\subsection{Related Datasets}
\label{datasets}
\textbf{Video Captioning:} TACoS \cite{regneri-etal-2013-grounding} offers fine-grained cooking activities, while the MSVD \cite{chen-dolan-2011-collecting}, MPII-MD \cite{rohrbach15cvpr}, and M-VAD \cite{pini2019mvad} datasets present a broad open domain with a substantial volume of clips. Although the MSR-VTT \cite{xu2016msr} dataset is rich in movie scene captions and provides a fundamental scene-based approach, it lacks the specificity required for fine-grained descriptions.
The Charades \cite{sigurdsson2016hollywood} and Charades-Ego \cite{sigurdsson2018actor} datasets contribute valuable insights into daily indoor activities with lengthy captions. The ActivityNet Captions \cite{caba2015activitynet} dataset broadens the domain of instance-based activities as a dense captioning task with a significant number of clips, but it does not match the level of detail in pedestrian behavior that WTS offers.
The large-scale instructional dataset HowTo100M \cite{miech19howto100m} encompasses a vast array of activities, but it provides limited length in caption information. TRECVID-VTT’20 \cite{awad2021trecvid} offers a noteworthy volume of open domain clips, yet it does not approach the intricacy of pedestrian-vehicle interactions as WTS does.
In the context of traffic-specific datasets, BDD-X \cite{kim2018textual} marks a significant step with its focus on traffic scenes and considerable annotation detail for driver action explanation. However, WTS surpasses it with higher granularity in pedestrian behavior analysis and a larger volume of clips and annotations focusing on pedestrians. Notably, WTS is pioneering in its inclusion of 3D gaze data, providing unparalleled insights into pedestrian attention and behavior in traffic scenarios.

\subsection{Video Captioning Methods}
Video and image captioning are fundamental tasks in video understanding. Vid2Seq \cite{yang2023vid2seq} introduces a model that integrates special time tokens in a language model to predict event boundaries and textual descriptions in the same sequence. T. Wang et al. \cite{wang2021end} present PDVC, a framework for dense video captioning that uses parallel decoding and treats dense caption generation as a set prediction task. MPLUG-2 \cite{Xu2023mPLUG2AM} leverages large-scale pre-training for a deep understanding of complex visual-language interactions. VALOR \cite{chen2023valor} is a framework for object retrieval tasks involving video and language input, excelling in processing complex queries and locating items based on descriptions.
Recent foundation models have significantly improved performance in video-to-text tasks due to prior knowledge alignment. DriveGPT-4 \cite{xu2023drivegpt4}, based on the GPT-4 architecture, integrates visual data and contextual understanding for autonomous driving scenarios, showing strong performance on the BDD-X \cite{kim2018textual} dataset. Caption Anything \cite{wang2023caption} generates accurate, context-aware captions for a wide range of video content, leveraging the segment anything model for descriptive and relevant captions. Video-LLaMA \cite{damonlpsg2023videollama} combines linguistic, visual, and audio data for comprehensive video content understanding.
We benchmarked recent Video LLM-based methods' performance on the WTS dataset to evaluate their potential for fine-grained video-to-text tasks.

\subsection{Evaluation Metrics}
Video/image caption evaluation metrics including, 
reference-based ones such as BLUE\cite{papineni-etal-2002-bleu}, ROUGE\cite{lin-2004-rouge}, CIDER\cite{oliveira-dos-santos-etal-2021-cider}, METEOR\cite{banerjee-lavie-2005-meteor}, SPICE\cite{anderson2016spice}
and Rankgen\cite{krishna-etal-2022-rankgen}. However, due to the above methods focused on the word level similarity and its order, 
for the long caption, especially its semantic meaning evaluation is relatively difficult to judge that two paragraphs represent the same thing but different words.
Recently, for video-language understanding benchmark, GPT-based metrics \cite{bai2023touchstone,fu2023gptscore,li2024mvbench} have been developed for use, which are better for aligning the semantic meaning for evaluation.
The main difference between our proposed LLMScorer and the above ones is 
LLMScore is designed with a customizable specified aspect considering the semantic meaning as well as syntactic structure similarity
for a holistic caption correctness evaluation.
\section{WTS Dataset}
\label{sec:dataset}

WTS is a novel pedestrian-centric traffic video dataset featuring $255$ traffic scenarios, including staged pedestrian-related accidents across $1.2k$ video segments. Each scenario spans $1$ to $3$ minutes, with segments ranging from $1$ to $15$ seconds. It covers 5 phases of pedestrian behavior (Pre-recognition, Recognition, Judgement, Action, Avoidance). Detailed textual descriptions of pedestrian and vehicle behaviors are provided for each segment, along with bounding box annotations. We also curated approximately $5k$ pedestrian-related videos from BDD100K \cite{Yu_2020_CVPR} using the same annotation approach as WTS. Additionally, we include synchronized 3D gaze and location annotations for each scenario video, totaling $52,823$ frames across $6$ subjects in outdoor environments.

\begin{figure}[t]
\centering
\includegraphics[scale=0.30]{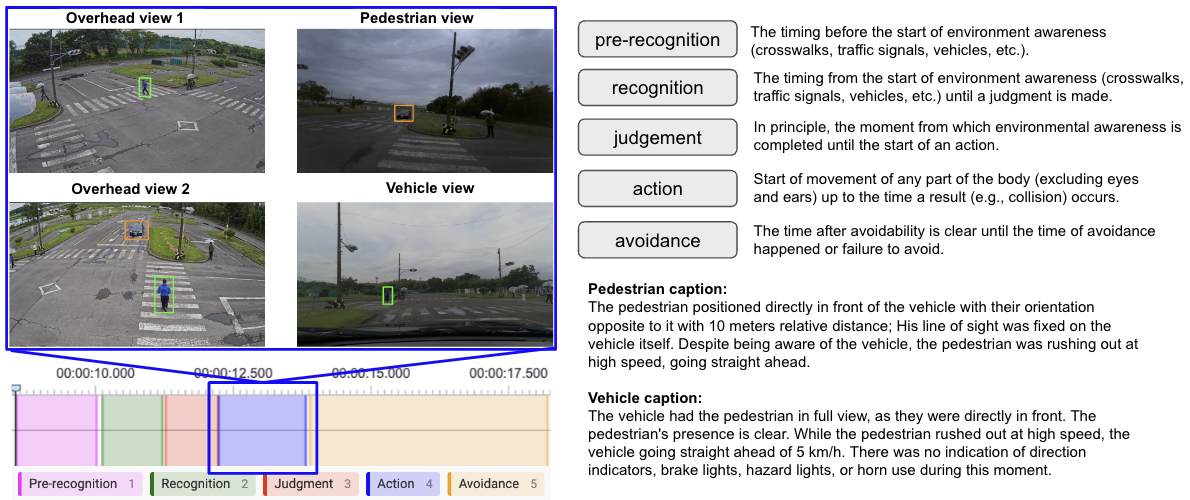}
   \caption{The overview of WTS video caption data structure: 
   1) the left figure shows multiple views from overhead to ego vehicle view with 5 phases.
   2) the right figure shows the definition of our phase segment and the GT captions corresponding with action segment about the target pedestrian and vehicle respectively as an example.}
\label{fig:caption_sample}
\end{figure}

\subsection{Data Construction}
\label{construction}

\textbf{Camera views:}
Our recording setup includes three types of cameras: overhead, driver recorder, and ego-centric cameras. The multi-view setup is designed for vehicle-infrastructure cooperation, such as in smart cities, enhancing the accuracy of fine-grained descriptions and improving AD system safety features. It also helps avoid false negatives from vehicle blind spots and offers promising avenues for future research.
We selected $18$ out of $24$ overhead views after removing occluded viewpoints. Each view records at $1080p$ resolution and $24$ fps, with calibrated camera parameters. The driver recorder uses a GoPro Hero10 with linear model settings at $1080p$ and $24$ fps. The ego-centric camera, Tobii Pro Glasses 3, captures $720p$ resolution videos at $24$ fps, providing accurate 2D gaze ground truth. Sample views of these cameras are illustrated in Figure \ref{fig:intro}.

\textbf{Subjects:}
There are total $14$ subjects who joined the recording with $7$ females and $7$ males, whose ages are ranged from $16{\sim}50$ years old.

\begin{figure}[t]
\centering
\includegraphics[scale=0.33]{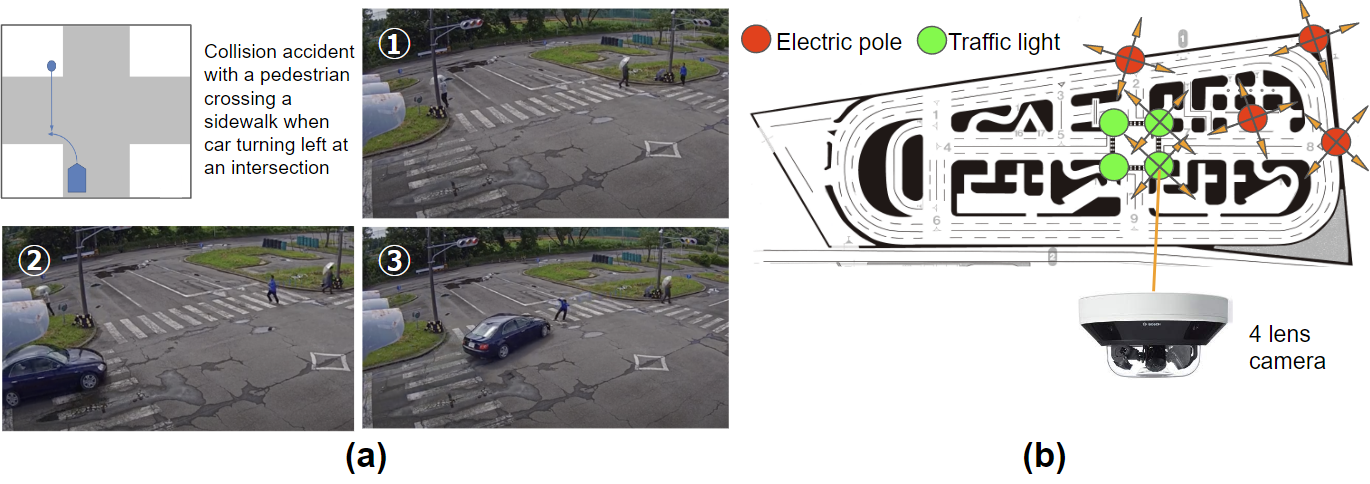}
   \caption{(a).Sample of scenario pattern. 
   3 frames are sampled from the video along the temporal direction with the order 1 to 3 at the upper left of the frame. (b).Our recording environment map and camera position.}
\label{fig:env_iso}
\end{figure}

\textbf{Scenarios:}
We follow the ISO34502 standard, a scenario-based safety evaluation framework used for automated driving systems as well as our scenario guideline.
We created at least one scenario for each of the $138$ pedestrian-vehicle relative position patterns defined in it to construct the recording scenario.
The scenario patterns and its recorded video sample frames are shown in Figure \ref{fig:env_iso}(a).

\textbf{Captions:}
We provide fine-grained pedestrian-related behavior captions for each video segment in the traffic scenarios.
It starts from the phase segmentation to find each behavior phase temporal localization part, then moves to describe the event in the segment into text along the temporal directions. 
The features of this caption can be drawn as 1. long paragraph; 2. fine-grained observation regarding the position, action, attributes, and attention of pedestrians and vehicles with the surrounding context; 3. focus on the target objective.

\textbf{3D Gaze:}
3D Gaze is provided for the target pedestrians as extra data for further use, such as using it as a prior for traffic accident analysis.
We provide both 2D and 3D gaze annotations for the corresponding frames with the target pedestrian. 
Figure \ref{fig:intro} shows the example of 2D/3D gaze annotations in the frame.
Except for the above-ground truth, we also provide the 3D head position and raw 2D gaze ground truth acquired from the Tobii glasses.
To check and visualize the 3D information appropriately, a 2D paired 3D scanned map is given. 

\subsection{Data Collection}
\label{collection}

Generally, collecting natural traffic events in the real world is challenge.
WTS ensures a controlled environment for collecting traffic events, adhering to ISO34502 scenarios and using pre-defined context settings based on actual accident videos. All accident-related scenarios are staged by professional stuntmen to replicate real-world conditions accurately.

\textbf{Recording environment:}
We use a driving school with several intersections and single roads as our recording environment.
It is a $72\times84$ meters outdoor space in total, and $15\times15$ meters for intersection and $11m$ as width and $81m$ as length for the single road.
We have installed 6 multi-lens (4 lenses) overhead view cameras \textit{BOSCH FLEXIDOME multi 7000i} in our recording place, and each overhead view camera is attached to the electric pole around the road. 
Figure \ref{fig:env_iso}(b) shows the map of our recording place with the camera placement positions.

\textbf{Recording process:}
A series of scenarios was listed on a worksheet. 
Random walk or standby action was performed by subjects before and after each event. 
Each event includes each phase from Pre-recognition to Avoidance.
The traffic light in the intersection is operated normally without any pre-defined behavior.
The events occurred at various positions to ensure the diversity of the dataset.
In each event, target pedestrians will be involved in the scenario, and non-target pedestrians who only performed the random walk in the video to make the task setting close to the real-world setup.

\textbf{Synchronization:}
Synchronizing multiple, heterogeneous videos is challenging, especially under unconstrained outdoor environments, where we have multiple dynamic cameras. 
We utilize the audio input that most commodity video cameras are equipped to synchronize the videos. We use analog radio signals as the audio syncing signals. 
The average sync delay is $0.015$ seconds, well below our frame interval of $24$ FPS. 
For quality assurance, human annotators checked for audio delays (echoing when two videos are out of sync by more than $0.03$ seconds) and labeled such videos for human modification.

\subsection{Ground Truth Generation}
\label{annotation}
Annotation of the detailed description of traffic video is not easy to ensure accuracy and bias from the human.
To resolve this kind of challenge and provide high-quality annotations for each data, we introduce several novel manners for the annotation.

\begin{figure}[t]
\centering
\includegraphics[scale=0.20]{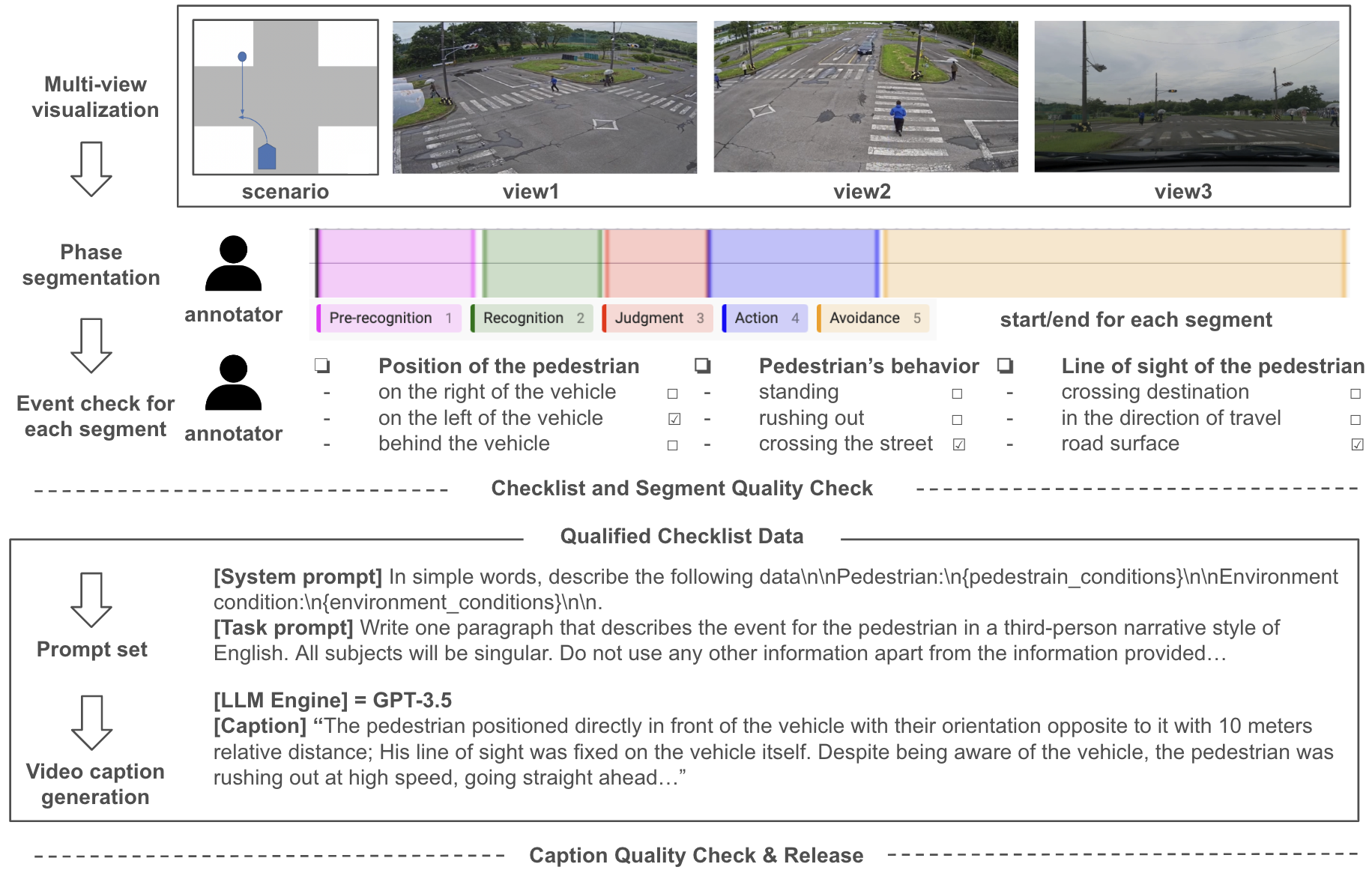}
   \caption{Annotation pipeline for generating the traffic domain-related captions.}
\label{fig:annotation_flow}
\end{figure}

\textbf{Captions:}
For the traffic event-related caption, two challenge points are: 
1) the caption information requires professional knowledge and viewpoint that is hard to create by a normal annotator correctly. 
2) to give the temporal localization information to each segment and create the related temporal-spatial behavior caption is still based on the intuition of the analyst's experience, which is biased.
3) writing a long description of the video from humans requires a high concentration which is error-prone.
To remove the bias and make the annotator perform the annotation without deep knowledge of the traffic safety analysis area, 
we introduced a semi-auto caption generation manner with the following flow as shown in Figure \ref{fig:annotation_flow}: 
1. We cooperated with the professional analyst in the insurance company to regularize and unify the guidelines for the segment temporal localization manner. 
Annotators are asked to do the phase segmentation according to the guidelines.
2. We make structured knowledge from the existing traffic event textual description data. The structured knowledge will be a breakdown of over 180 factual items as a checklist related to the environment context, attributes, position, action, and attention from target pedestrians and vehicles. For the position (left, right, front, etc.) items regarding the pedestrian and vehicle, we use the vehicle-centric as the relative anchor to define whether the pedestrian is "left" or right" to the vehicle to remove the bias. Samples of our checklist are shown in Figure \ref{fig:annotation_flow}. Details can be referred from the supplementary.
3. Annotators are asked to check the items from the structured checklist for each annotated segment according to facts that occurred in the video.
4. Once the event check process is done, a double-check process happens to verify whether the checked items match the video or not as the first quality verification.
5. Then, the checked items are fed to a Large Language Model with an appropriate prompt setting (detailed can be referred from the supplementary) for generating the natural sentence, including all the checked items as the caption ground truth. 
We use GPT-3.5 \cite{gpt3.5} as the LLM engine for the caption generation.
6. Finally, a double-check for the generated caption is done manually as the second quality verification.

Notice that GPT is used solely to summarize human-annotated scenario-describing checklist results into captions, ensuring that the diversity of scenario descriptions is not limited by GPT's diversity.

\begin{figure}[t]
\centering
\includegraphics[scale=0.77]{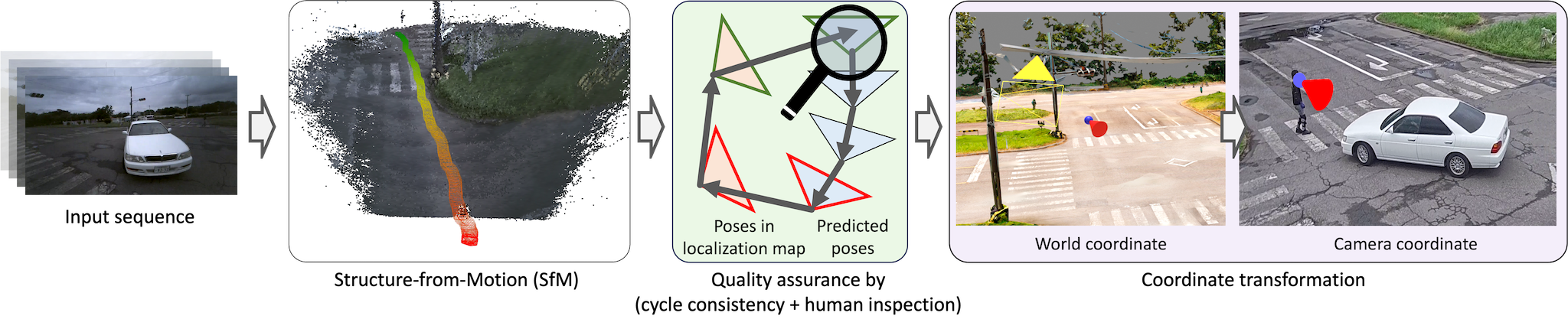}
   \caption{Illustration of the 3D gaze GT pipeline. Note that the radius of the red cone, illustrating the 3D gaze direction, is solely for visualization purposes. It represents an approximate eccentric gaze FOV of 15\textdegree, attainable without head movement.}
\label{fig:gaze_pipeline}
\end{figure}

\begin{figure}[t]
\centering
\includegraphics[scale=0.30]{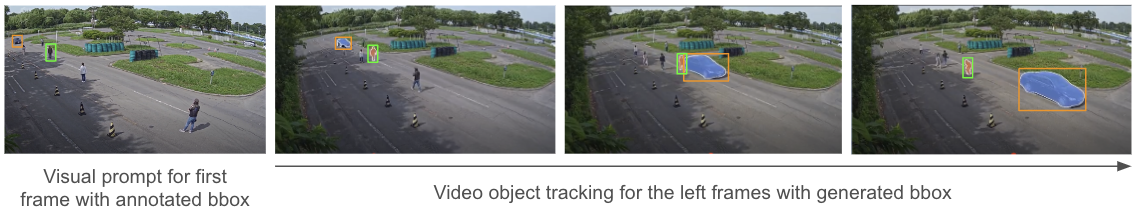}
   \caption{Qualitative sample from our tracking results for bbox generation. Using human annotated bbox as visual prompt input to the track-anything.}
\label{fig:tracking_bbox}
\end{figure}

\textbf{3D Gaze:}
Previous studies on 3D gaze ground truth have typically been conducted in controlled environments or with people under controlled conditions. These studies often rely on numerous AR markers to localize ego-view cameras \cite{nonaka2022dynamic} or restrict the point of gaze, providing instructions to subjects for gaze estimation \cite{gaze360_2019,hu2023gfie}. However, these methods can make the environment appear unnatural in third-person view videos, or hinder subjects from acting naturally.

To overcome these limitations, our approach involves localizing in the environment through structure-from-motion (SfM) from ego-view video, inspired by \cite{HPS}. The pipeline is outlined in Figure \ref{fig:gaze_pipeline}. Input videos are sub-sampled at 5 fps, and each frame is localized in world coordinates using SfM and a pre-built localization map. We utilize the GTSfM open source SfM library \cite{GTSFM} and a Matterport camera with LiDAR scanning \footnote{https://matterport.com/} to create the pre-built localization map.
Finally, the 3D gaze direction in ego-view is transformed into each surveillance camera's third-person view. This is done using the pre-calibrated surveillance camera pose, the ego-view frame pose, and the 3D gaze direction from the Tobii glasses in local coordinates. The transformation process for the $i$-th surveillance camera is defined as:$\mathbf{d}_i = \mathbf{R}_i^{-1}\mathbf{R}_{\text{ego}}\mathbf{d}_{\text{ego}}$,
where $\mathbf{d}_{\text{ego}}\in \mathbb{S}^2$ represents the 3D gaze direction in ego-view, $\mathbf{R}_{\text{ego}}\in \text{SO}(3)$ the ego-view pose in world coordinates, $\mathbf{R}_i\in \text{SO}(3)$ the $i$-th surveillance camera pose in world coordinates, and $\mathbf{d}_i \in \mathbb{S}^2$ the 3D gaze direction in the $i$-th surveillance camera. 

We evaluated the 3D gaze annotation quality in four aspects: 
(1) ego-view pose estimation using SfM, (2) transformation to world coordinates, (3) overhead camera pose estimation, and (4) Tobii’s accuracy with moving subjects.
For (1), sampled videos with rapid motion were used, resulting in an average error of $4.19$ degrees. 
For (2) and (3), we estimated relative poses between real and rendered images from scanned 3D scenes, achieving an error of $0.18$ degrees. 
We applied a sanity check to remove visually perceptible errors and aggregated ego-view poses using RANSAC and Procrustes. 
For (4), Tobii’s gaze accuracy for walking subjects was $1.74$ degrees \cite{tobii}. 
Finally, the combined error was $6.11$ degrees, within the human eye's $15$ degree eccentric gaze FOV.

\textbf{Bounding Box Generation}
We also provide bounding boxes of target vehicles and pedestrians. We choose a semi-supervised approach based on the human prompt in the first frame and tracking the target in the rest of the frames. We leverage Track Anything \cite{TrackAnything}, an interactive tool for segmentation and video object tracking based on Segment Anything \cite{SAM} and XMem \cite{XMem} respectively, which only takes several clicks on the target in the first frame as input.


\section{Baseline Approach}
\label{sec:baseline_approach}

Based on our dataset, we prepared three baselines for testing the fine-grained video captioning task.
\textbf{Video-LLaMA}\cite{zhang2023videollama} is a multi-modal LLM framework with the capability of understanding both visual and auditory content in the video. The video-language branch is composed of a frozen pre-trained image encoder from EVA-CLIP ViT-G/14 to extract features from video frames In our experiment, we do not use the audio branch but only the pre-trained video-language branch for the video caption benchmark without fine-tuning on the WTS dataset under several different prompt settings.
\textbf{Video-ChatGPT}\cite{Maaz2023VideoChatGPT} use CLIP ViT-L/14 as the visual encoder. To acquire the video-level feature, it uses frame-level embeddings are average-pooled along the temporal dimension to obtain a video-level temporal representation. Similarly, the frame-level embeddings are average-pooled along the spatial dimension to yield
the video-level spatial representation. The temporal and spatial features are concatenated to obtain the video-level features with a linear layer to project the video embedding $Q_v$ into the language decoder’s embedding space. The text queries are tokenized to the same dimensions as $Q_t$ concatenated with the $Q_v$ input to the language decoders.  

\begin{figure}
\centering
\includegraphics[scale=0.22]{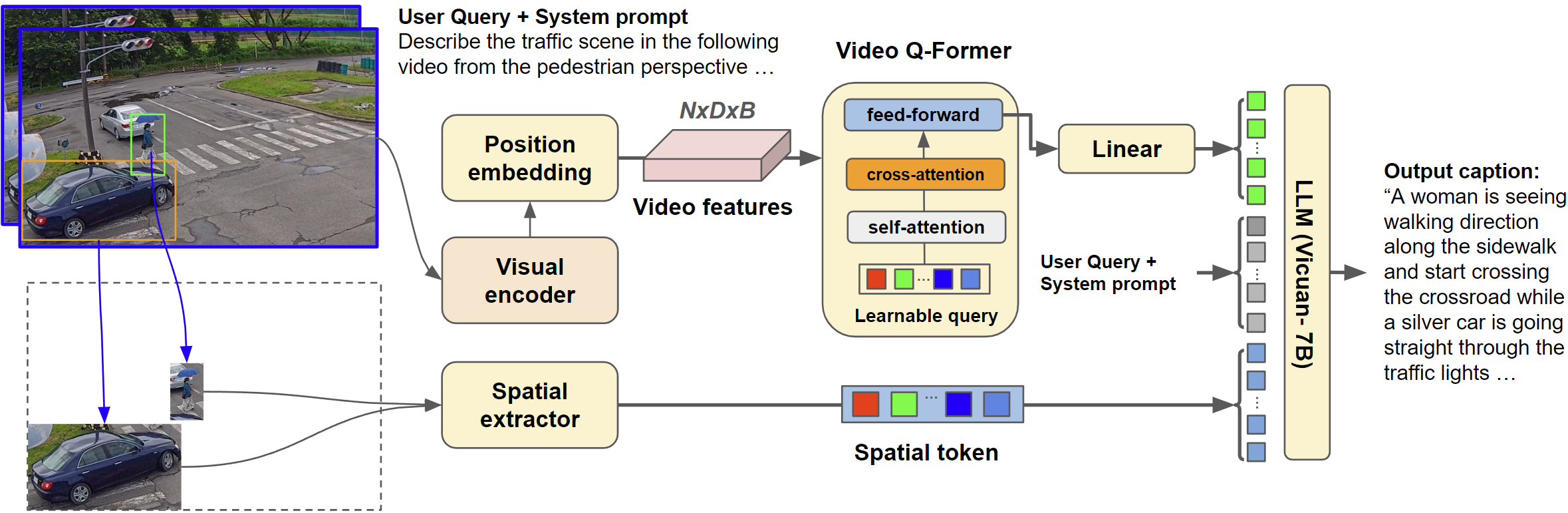}
   \caption{Instance-VideoLLM approach overview.}
\label{fig:approach}
\end{figure}

\textbf{Instance-VideoLLM} is our proposed baseline with fine-tuning on our training dataset. The main framework follows a similar architecture to Video-LLaMA, thus we use the same visual encoder, positional embedding, and Video Q-former as Video-LLLaMA. As the caption is targeted at the specific pedestrian or vehicle, we introduce a spatial token that represents the target instance region to the language encoder. The overview of the approach is shown in Figure \ref{fig:approach}.
For a given video $V$, the segments of the video $M$ is defined as $M_{seg}$. The frame $F_t, t\in N$ from $M_{seg}$ is fed to the visual encoder and then applied with a positional embedding to the representations of different frames. The position-encoded frame representations with dimension $D$ and frame number $N$ in a batch size $B$ are denoted with tensor ${N\times D\times B}$ are fed to the Video Q-Former to obtain the video embeddings. A linear layer to transform the video embeddings into the video query vectors $V$.
The video query vectors are of the same dimension as the text embeddings $T$ of LLM. 
Inspired by Pixel LLM\cite{yuan2023osprey} using the binary mask to introduce the instance spatial information. Each video segment's first frame target bounding box region will be used for generating a bbox mask binary map, which will be resized into $224\times224$ fulfilled with a binary mask as 1 if the pixel in the bbox. Then flatten and project it to generate the spatial token $S$. In the forward pass, video query embedding $V$ and spatial token $S$ will be concatenated to text embeddings $T$ as a video-instance soft prompt to guide the frozen LLMs to generate the text conditioned on video content

\section{Experiment}
\label{sec:experiment}



\def\arraystretch{1.5}
\begin{table}[t]
\centering
\begin{adjustbox}{width=\textwidth}
\begin{tabular}
{l c c c | c c c c}
\hline
Method & Prompt & LLM & Fine-tune & BLUE-4 & METEOR & ROUGE-L & CIDER\\
\hline
Video-LLaMA \cite{zhang2023videollama} & P-A & LLaVA-7B & No &0.022 &0.201 &0.195 &0.119 \\
Video-ChatGPT \cite{Maaz2023VideoChatGPT} & P-A & Vicuna-7B & No &0.096 &0.117 &0.171 &0.009 \\
\hline
Video-LLaMA \cite{zhang2023videollama} & P-B & LLaVA-7B & No &0.027 &0.210 &0.211 &0.143 \\
Video-ChatGPT \cite{Maaz2023VideoChatGPT} & P-B & Vicuna-7B & No &0.024 &0.178 &0.208 &0.053 \\
\hline
Video-LLaMA \cite{zhang2023videollama} & P-C & LLaVA-7B & No &0.045 &0.247 &0.226 &0.210 \\
Video-ChatGPT \cite{Maaz2023VideoChatGPT} & P-C & Vicuna-7B & No &0.072 &0.267 &0.266 &0.282 \\
\hline
Ours(VideoLLM) & P-C & Vicuna-7B & YES &0.101 &0.389 &0.407 &0.363 \\
Ours(Instance-VideoLLM) & P-C & Vicuna-7B & YES &0.121 &0.409 &0.417 &0.389 \\
\hline
\end{tabular}
\end{adjustbox}
\caption{Average performance comparison on the WTS dataset. Both WTS staged data and BDD are used for evaluation.}
\label{tbl:all_result}
\end{table}

\def\arraystretch{1.2}
\begin{table}[t]
\centering
\begin{adjustbox}{width=\textwidth}
\begin{tabular}
{l c | c c c}
\hline
Method  & Fine-tune & Semantic & Syntactic & LLMScore \\
\hline
Video-LLaMA \cite{zhang2023videollama} & No &0.008 &0.373 &0.117 \\
Video-ChatGPT \cite{Maaz2023VideoChatGPT} & No &0.004 &0.468 &0.143 \\
\hline
Ours(Instance-VideoLLM) & YES &0.285 &0.508 &0.351 \\
\hline
\end{tabular}
\end{adjustbox}
\caption{Comparison methods on LLMScorer metric.}
\label{tbl:llmscore_table}
\end{table}

\begin{figure}[t]
\centering
\includegraphics[scale=0.30]{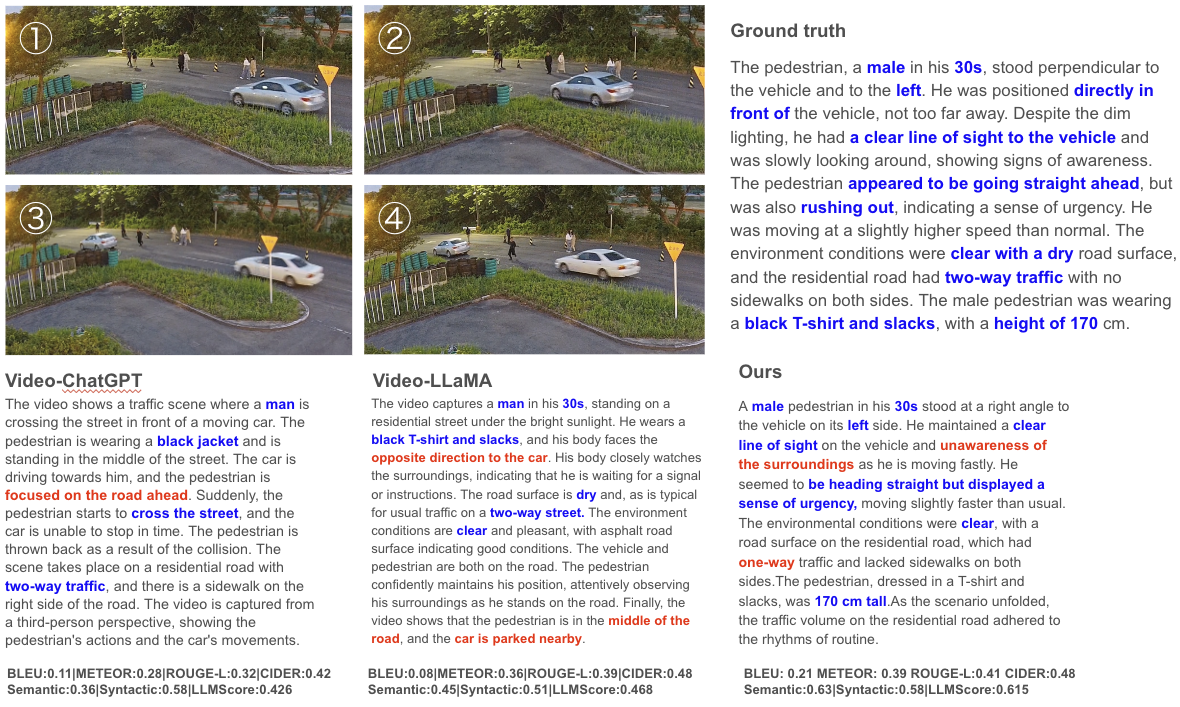}
   \caption{Comparison of the caption from each method evaluated with each metric. 
   Blue font means the evaluation target aspect in GT and correct representation, and Red font means the error representation.}
\label{fig:case}
\end{figure}

\subsection{Experiment Setup}
{\textbf{Dataset for benchmarking:}}
There are around 120 scenarios from staged data and 2000 scenarios from selected BDD data as training sets, 60 scenarios from staged, and 800 scenarios from BDD as validation sets. Each scenario will have $\sim5$ segments with $\sim5$ captions. For staged data, as there are multiple overhead views, we picked up a main view that covers the whole scenario in one view with a clear visible angle for the pedestrian and vehicle from the dataset for training and validation purposes. We think multiple-view caption consistency is also a new aspect for pushing forward more accurate video-to-text performance.    

{\textbf{LLMScore Evaluation protocol:}}
Semantic and syntactic information is crucial to measure the relatedness of two sentences. Various studies \cite{huang2021disentangling} \cite{Fei2020ImprovingTU} learn to disentangle the semantic and syntactic representations.
To achieve this, inspired by the evaluation protocol in GPTScore \cite{fu2023gptscore}, our LLMScore has a prompt template includes task description (comparing two captions), ground truth caption ($G$), inferred caption ($C$), and consideration aspects (location, attention, behavior of pedestrian/vehicle, and environment). 
\textbf{Semantic score} quantifies the degree of semantic similarity between two sentences. In our approach, we instruct the LLM (\textit{GPT-3.5-turbo}) to evaluate the semantic accuracy of the caption for each aspect \textit{Location, Attention, Behavior, Context} with respect to the ground truth. We assign a score of $1$ to the aspect that is semantically correct, otherwise $0$. We take an equal-weighted average of these scores and call this Semantic Score ($Score_{sem}$).
\textbf{Syntactic score} quantifies the syntactic similarity between the answers. First, for each aspect, we prompt LLM (\textit{GPT-3.5-turbo}) to give the answers to subjective questions for candidate caption as well as ground truth. Subsequently, we compute the cosine similarity score between the embeddings of these answers for the subject questions associated with candidate caption and ground truth. We use OpenAI's \textit{text-embedding-3-small} for generating embeddings. We then calculate the syntactic score ($Score_{syn}$) by taking the equally weighted average of the cosine similarities. Semantic score and Syntactic score are defined as $Score_{sem}$ and $Score_{syn}$ respectively as,

\begin{equation}
Score_{sem} = \frac{1}{n} \cdot \sum_{i=1}^{n} P_{sem}(T, A, G, C, Q_{sem}, O_{sem}, S_{sem})
\end{equation}

\begin{equation}
Score_{syn} = \frac{1}{n} \cdot \sum_{i=1}^{n} P_{syn}(T, A, G, C, Q_{syn}, O_{syn}, S_{syn})
\end{equation}

where $P_{sem}$ and $P_{syn}$ are the prompt template with the input, $T$ is task definition, $A$ is definition for aspects, $G$ is ground truth, $C$ is inferred caption. $Q_{sem}$ defines the queries, $S_{sem}$ is scoring criteria, $O_{sem}$ is output format for semantic scoring and $Q_{syn}$ defines the queries, $S_{syn}$ is scoring criteria, $O_{syn}$ is output format for syntactic scoring.
Therefore, LLMScore is defined as $LLMScore = w_1 * Score_{sem} + w_2 * Score_{syn}$, where $w_1$ and $w_2$ are the weights for the scores. 

\subsection{Implementation details}
For the Video-LLaMA and Video-ChatGPT, we did not fine-tune the model but used the input video and the user query prompt fed to the LLM for generating the captions.
There are three kinds of prompts we used, \textit{P-A} is a simple task prompt like "Describe the video from a pedestrian perspective". \textit{P-B} is a prompt with system settings and more constraints for the traffic domain for the task. \textit{P-C} is a system prompt with a task description following a demonstration sample. More detail can be referred from the supplementary. For our proposed baseline, during the fine-tuning, LLM is frozen only to fine-tune the Q-Former, linear translation part. We use the AdamW optimizer and a weight decay of $0.05$. 
We use a cosine learning rate decay with a peak learning rate of $2e-5$ and a linear warmup of $2k$ steps. We use images of size $22\times224$. For LLMScore, the $w_1$ and $w_2$ are set to $0.7$ and $0.3$ respectively. 

\subsection{Evaluations}
Table \ref{tbl:all_result} shows the comparison result on the WTS dataset for each method with different prompt settings on 4 kinds of traditional popular metrics. It is obvious that the prompt $P-C$ is the best setup and thus all the methods could achieve the best results under this setting. However, based on this best prompt setting, Video-LLaMA and Video-ChatGPT still worked not well for the WTS dataset traffic event domain showing that the fine-grained description in WTS is not generalizable from the common sense knowledge trained Video LLM model without fine-tuning. 

To evaluate the impact of fine-tuning on the WTS dataset, we compared \textit{Ours(VideoLLM)} to Video-LLaMA in Table 2 of our paper. Both models share similar architectures, isolating the effect of fine-tuning. The comparison showed that fine-tuning improves performance but is still a challenge. 
In \textit{Ours(Instance-VideoLLM)}, we added region-specific information. Despite this, results indicate significant room for improvement, suggesting our approach as an initial idea for developing more advanced methods for fine-grained instance-level video understanding with LLMs.

Table \ref{tbl:llmscore_table} shows the result using our LLMScore for each method. For Video-LLaMA and Video-ChatGPT, the semantic score is relatively low and the syntactic score is high, is because almost all the critical semantic meaning regarding the location, attention, behavior, and context are not correct according to the GT even though the whole paragraph looks similar to each other. More samples can be referred from the supplementary.
Figure \ref{fig:case} shows a success case sample of how the caption looks like from each method. It is hard to tell the difference from the traditional metrics but highly be separated by using the LLMScore for fine-tuned results, as more semantic meanings are correct for this case.  

To compare the LLMScore with the human evaluation result, We use 50\% of the validation set for human evaluations. Human evaluators score the correctness of information in $C$ and $G$ using pre-set questions, extracting sub-texts that best describe the aspects. The average human score is $0.243$ (variance is $0.002$), close to the LLMScore of $0.242$ (variance is $0.0007$) for the same samples.
\section{Conclusion}
\label{sec:conclusion}
We introduced the WTS dataset a large-scale pedestrian-centric traffic dataset accompanied by detailed textual descriptions of both vehicle and pedestrian behavior and 3D gaze meta-information for pedestrian use. A new LLM-based video caption evaluation scorer and an Instace-VideoLLM baseline are proposed as well. WTS is a challenging dataset with long descriptions for the traffic video domain, experiment shows that there is a large space for pushing forward the spatial-temporal language understanding into the next stage.

%
%
\bibliographystyle{splncs04}
\bibliography{main}
\end{document}